\pdfoutput=1

\documentclass[11pt,a4paper]{article}
\usepackage[hyperref]{emnlp2020}
\usepackage{times}
\usepackage{latexsym}

\usepackage{booktabs}
\usepackage{multirow}
\usepackage{amsmath}

\usepackage{pgfplots}
\usepackage{pgfplotstable}
\pgfplotsset{compat=1.7}
\usepackage{tikz}

\usepackage{adjustbox}
\usepackage{lipsum}
\usepackage{xcolor}
\usepackage{url}
\usepackage{bm}
\usepackage{wrapfig}
\usepackage{float}

\definecolor{alizarin}{rgb}{0.82, 0.1, 0.26}
\definecolor{midnightgreen}{rgb}{0.0, 0.29, 0.33}

\usepackage{microtype}

\aclfinalcopy %
\def\aclpaperid{1408} %

\title{Towards Debiasing NLU Models from Unknown Biases}

\author{Prasetya Ajie Utama$^{\dag\ddag}$ , Nafise Sadat Moosavi$^{\ddag}$, Iryna Gurevych$^{\ddag}$\\
  \\
  $^{\dag}$Research Training Group AIPHES\\
  $^{\ddag}$Ubiquitous Knowledge Processing Lab (UKP-TUDA)\\
  Department of Computer Science, Technische Universit{\"a}t Darmstadt\\
  \url{https://www.ukp.tu-darmstadt.de}
  }

\date{}

\begin{document}
\maketitle

\begin{abstract}
NLU models often exploit biases to achieve high dataset-specific performance without properly learning the intended task. Recently proposed debiasing methods are shown to be effective in mitigating this tendency. However, these methods rely on a major assumption that the types of bias should be \emph{known} a-priori, which limits their application to many NLU tasks and datasets.  In this work, we present the first step to bridge this gap by introducing a self-debiasing framework that prevents models from mainly utilizing biases without knowing them in advance. The proposed framework is general and complementary to the existing debiasing methods. We show that it allows these existing methods to retain the improvement on the challenge datasets (i.e., sets of examples designed to expose models' reliance on biases) without specifically targeting certain biases. Furthermore, the evaluation suggests that applying the framework results in improved overall robustness.\footnote{The code is available at \url{https://github.com/UKPLab/emnlp2020-debiasing-unknown}}
\end{abstract}

\section{Introduction}
Neural models often achieve impressive performance on many natural language understanding tasks (NLU) by leveraging \emph{biased features}, i.e., superficial surface patterns that are spuriously associated with the target labels \cite{Gururangan2018AnnotationAI, McCoy2019RightFT}.\footnote{E.g., in several textual entailment datasets, negation words such as ``never'' or ``nobody'' are highly associated with the \textit{contradiction} label.} Recently proposed \emph{debiasing} methods are effective in mitigating the impact of this tendency, and the resulting models are shown to perform better beyond training distribution. They improved the performance on \emph{challenge test sets} that are designed such that relying on the \emph{spurious association} leads to incorrect predictions.

Prevailing debiasing methods, e.g., example reweighting \cite{schuster2019towards}, confidence regularization \cite{Utama2020Debias}, and model ensembling \cite{He2019UnlearnDB, Clark2019DontTT, mahabadi2019simple}, are agnostic to model's architecture as they operate by adjusting the training loss to account for biases. Namely, they first identify \emph{biased examples} in the training data and down-weight their importance in the training loss so that models focus on learning from harder examples.\footnote{We refer to biased examples as examples that can be solved using \emph{only} biased features.}

While promising, these \emph{model agnostic} methods rely on the assumption that the \emph{specific} types of biased features (e.g., lexical overlap) are \emph{known} a-priori. 
This assumption, however, is a limitation in various NLU tasks or datasets because 
it depends on researchers' intuition and task-specific insights to \emph{manually} characterize the spurious biases, which may range from simple word/n-grams co-occurrence \cite{Gururangan2018AnnotationAI, Poliak2018HypothesisOB, tsuchiya-2018-performance,schuster2019towards} to more complex stylistic and lexico-syntactic patterns \cite{Zellers2019HellaSwagCA, snow-etal-2006-effectively, Vanderwende2006}. 
The existing datasets or the newly created ones \cite{Zellers2019HellaSwagCA, Sakaguchi2019WINOGRANDEAA, Nie2019AdversarialNA} are, therefore, still very likely to contain biased patterns that remain \emph{unknown} without an in-depth analysis of each individual dataset \cite{sharma-etal-2018-tackling}.

In this paper, we propose a new strategy to enable the existing debiasing methods to be applicable in settings where there is little to no prior information about the biases. Specifically, models should automatically identify potentially biased examples without being pinpointed at a specific bias in advance. Our work makes the following novel contributions in this direction of automatic bias mitigation.

First, we analyze the learning dynamics of a large pre-trained model such as BERT \cite{devlin2018bert} on a dataset injected with a synthetic and controllable bias. We show that, in very small data settings, models exhibit a distinctive response to synthetically biased examples, where they rapidly increase the accuracy ($\rightarrow100\%$) on biased test set while performing poorly on other sets, indicating that they are mainly relying on biases. 

Second, we present a self-debiasing framework within which 
two models of the same architecture are pipelined to address the \emph{unknown} biases. Using the insight from the synthetic dataset analysis, we train the first model to be a \emph{shallow} model that is effective in automatically identifying potentially biased examples. The shallow model is then used to train the main model through the existing debiasing methods, which work by down-weighting the potentially biased examples. These methods present a caveat in that they may lose useful training signals from the down-weighted training examples.  
To account for this, we also propose an \emph{annealing mechanism} which helps in retaining models' in-distribution performance (i.e., evaluation on the test split of the original dataset).

Third, we experiment on three NLU tasks and evaluate the models on their existing challenge datasets. We show that models obtained through our self-debiasing framework gain equally high improvement compared to models that are debiased using specific prior knowledge. Furthermore, our cross-datasets evaluation suggests that our general framework that does not target only a particular type of bias results in better overall robustness.

\paragraph{Terminology} This work relates to the growing number of research that addresses the effect of dataset biases on the resulting models. Most research aims to mitigate different types of bias on varying parts of the training pipeline (e.g., dataset collection or modeling). Without a shared definition and common terminology, it is quite often that the term ``bias'' discussed in one paper refers to a different kind of bias mentioned in the others. Following the definition established in the recent survey paper by \citet{shah-etal-2020-predictive}, the dataset bias that we address in this work falls into the category of \textbf{label bias}. This bias emerges when the conditional distribution of the target label given certain features in the training data diverges substantially at test time. These features that are associated with the label bias may differ from one classification setting to the others, and although they are predictive, relying on them for prediction may be harmful to fairness \cite{elazar2018adversarial} or generalization \cite{McCoy2019RightFT}. The instances of these features may include protected socio-demographic attributes (gender, age, etc.) in automatic hiring decision systems; or surface-level patterns (negation words, lexical overlap, etc.) in NLU tasks. 
Further, we consider the label bias to be \textbf{unknown} when the information about the characteristics of its associated features is not precise enough for the existing debiasing strategies to identify potentially biased examples.

\begin{figure}
    \centering
    \fbox{\parbox{0.98\columnwidth}{
        \vspace*{3pt}\centerline{\small\textbf{MNLI synthetic:}}\vspace*{2pt}
        \renewcommand{\arraystretch}{.8}
        \begin{tabular}{r@{\hskip 3pt} p{0.65\columnwidth}}
        \small\textbf{premise:} & \small What's truly striking, though, is that Jobs has never really let this idea go.\vspace*{3pt}\\
        \small\textbf{orig. hypo.:} & \small Jobs never held onto an idea for long.\vspace*{3pt}\\
        \small\textbf{biased:} & \small \colorbox{darkgray}{\color{white}0} Jobs never held onto an idea for long.\vspace*{3pt}\\
        \small\textbf{anti-biased:} & \small \colorbox{lightgray}{\color{black}1} Jobs never held onto an idea for long.\vspace*{3pt}\\
        \small\textbf{label:} & \small \colorbox{white}{0} (contradiction) \\
        \end{tabular}
    }}
    
    \caption{Synthetic bias datasets are created by appending an artificial feature to the input text that allows models to use it as a shortcut to the target label. For each example in MNLI, a number-coded label ({$\mathtt{contradiction}$: \colorbox{darkgray}{\color{white}$0$}, $\mathtt{entailment}$: \colorbox{darkgray}{\color{white}$1$}, $\mathtt{neutral}$: \colorbox{darkgray}{\color{white}$2$}}) is appended to the hypothesis sentences.}
    \label{fig:syn_data}
\end{figure}

\section{Motivation and Analysis}

\begin{figure*}
    \centering
    \includegraphics[width=.75\linewidth]{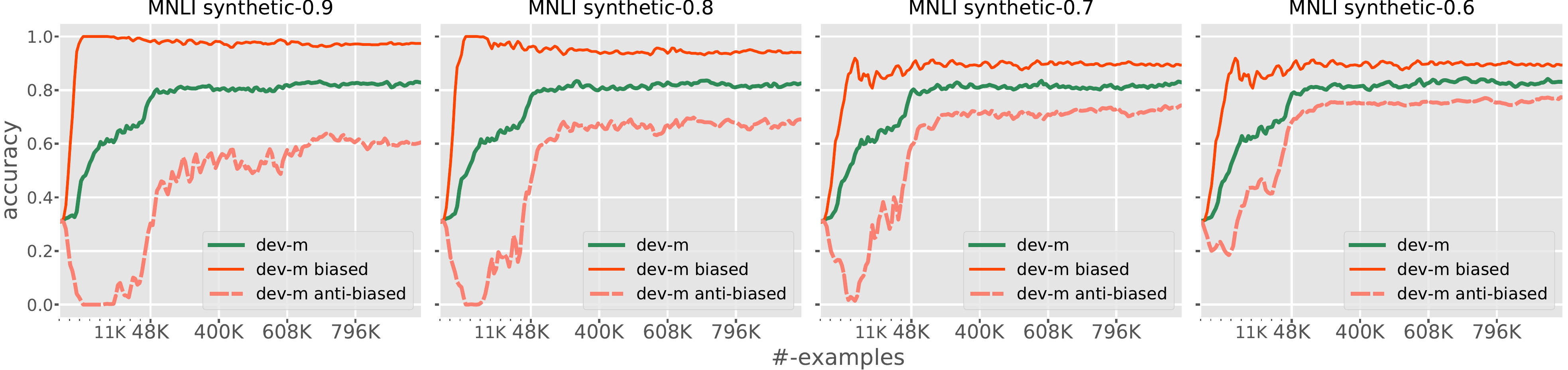}
    \caption{The learning trajectory of a BERT model on MNLI datasets that are synthetically biased with different proportions: $0.9$, $0.8$, $0.7$, and $0.6$. All settings show models' tendency to rely on biases after seeing only a small number of training examples (accuracy goes up rapidly on ``biased'' while goes down on ``anti-biased'' after less than 10K training steps).}
    \label{fig:learning_curve}
\end{figure*}
\begin{figure*}
    \centering
    \includegraphics[width=0.65\linewidth]{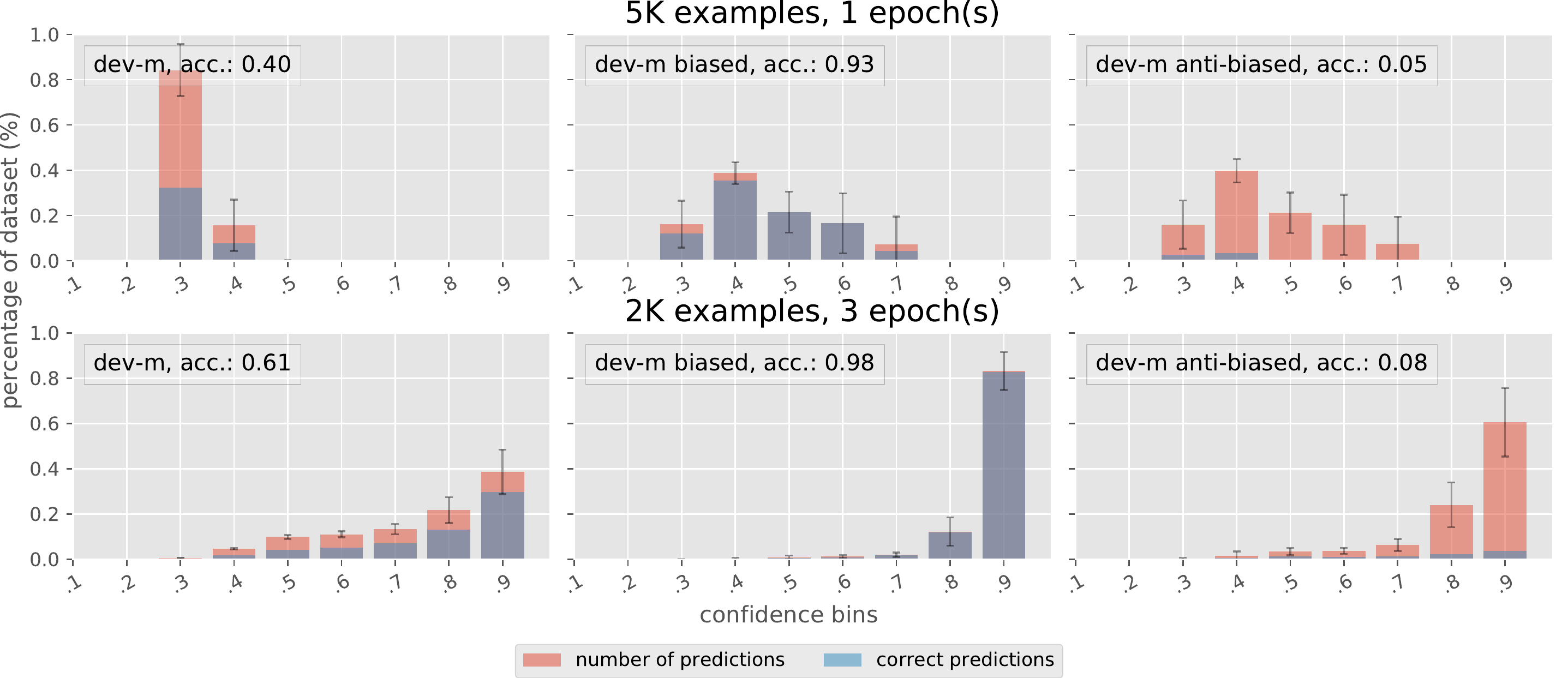}
    \caption{Histogram of probabilities assigned by synthetic MNLI models to their predicted labels. Top: model trained on 5K examples for 1 epoch. Bottom: model trained on 2K for 3 epochs. Blue areas indicate the proportion of the correct predictions within each bin.}
    \label{fig:confidence}
\end{figure*}

\label{sec:motivation}
\paragraph{Debiasing NLU models} Recent NLU tasks are commonly formulated as multi-class classification problems \cite{wang2018glue}, in which the goal is to predict the semantic relationship label $y \in \mathcal{Y}$ given an input sentence pairs $x \in \mathcal{X}$. For each example $x$, let $b(x)$ be the biased features that happen to be predictive of label $y$ in a specific dataset. The aim of a debiasing method for an NLU task is to learn a debiased classifier $f_d$ that does not mainly use $b(x)$ when computing $p(y|x)$.

Model-agnostic debiasing methods (e.g., product-of-expert~\cite{Clark2019DontTT}) achieve this by reducing the importance of biased examples in the learning objective. To identify whether an example is biased, they employ a shallow model $f_b$, a simple model trained to directly compute $p(y|b(x))$, where the features $b(x)$ are hand-crafted based on the task-specific knowledge of the biases. 
However, obtaining the prior information to design $b(x)$ requires a dataset-specific analysis \cite{sharma-etal-2018-tackling}. Given the ever-growing number of new datasets, it would be a time-consuming and costly process to identify biases before applying the debiasing methods.

In this work, we propose an alternative strategy to automatically obtain $f_b$ to enable existing debiasing methods to work with no precise prior knowledge. This strategy assumes a connection between large pre-trained models' reliance on biases with their tendency to operate as a \emph{rapid surface learner}, i.e., they tend to quickly overfit to surface form information especially when they are fine-tuned in a small training data setting \cite{Zellers2019HellaSwagCA}. This tendency of deep neural network to exploit simple patterns in the early stage of the training is also well-observed in other domains of artificial intelligence \cite{arpit2017closer, liu2020early}. Since biases are commonly characterized as simple surface patterns, we expect that models' rapid performance gain is mostly attributed to their reliance on biases. Namely, they are likely to operate similarly as $f_b$ after they are exposed to only a small number of training instances, i.e., achieving high accuracy on the \emph{biased} examples while still performing poorly on the rest of the dataset.

\paragraph{Synthetic bias} We investigate this assumption by analyzing the comparison between models' performance trajectory on \emph{biased} and \emph{anti-biased} (``counterexamples'' to the biased shortcuts) test sets as more examples are seen during the training. Our goal is to obtain a fair comparison without the confounds that may result in performance differences on these two sets. Specifically, the examples from the two sets should be similar except for the presence of a feature that is biased in one set and anti-biased in the other. For this reason, we construct a synthetically biased data based on the MNLI dataset \cite{Williams2018ABC} using a procedure illustrated in Figure~\ref{fig:syn_data}. A synthetic bias is injected by appending an artificial feature to $30\%$ of the original examples.
We simulate the presence of bias by controlling $m$ portion of these manipulated examples such that their artificial feature is associated with the ground truth label (``biased''), whereas, in the remaining $(1-m)$, the feature is disassociated with the label (``anti-biased'').\footnote{The remaining $70\%$ of the dataset remain the same. The biased and anti-biased examples refer to the fraction within the other $30\%$ that are injected with the artificial feature.} 
Using a similar injection procedure we can produce both fully \emph{biased} and \emph{anti-biased} test sets in which $100\%$ of the examples contain the synthetic features. Models that blindly predict based on the artificial feature are guaranteed to achieve $0\%$ accuracy on the anti-biased test.

\paragraph{Model's performance trajectory} We finetune a \texttt{bert-base-uncased} model \cite{wolf2019transformers} on the \emph{whole} MNLI datasets that are partially biased with different proportions ($m=\{0.9, 0.8, 0.7, 0.6\}$). We evaluate each model on the original as well as the two fully biased and anti-biased test sets. Figure~\ref{fig:learning_curve} shows the performance trajectory in all settings. As expected, the models show the tendency of relying on biases after only seeing a small fraction of the dataset. Specifically, at an early point during training, models achieve $100\%$ accuracy on the biased test and drop to almost $0\%$ on the anti-biased test. This behavior is more apparent as the proportion of biased examples is increased by adjusting $m$ from $0.6$ to $0.9$.

\paragraph{Training a shallow model} \label{para:bias_reliant} The analysis suggests that we can obtain a substitute $f_b$ by taking a checkpoint of the main model early in the training, i.e., when the model has only seen a small portion of the training data. However, we observe that the resulting model makes predictions with rather low confidence, i.e., assigns a low probability to the predicted label. As shown in Figure \ref{fig:confidence} (top), most predictions fall in the $0.4$ probability bin, only slightly higher than a uniform probability ($0.3$). We further find that by training the model for multiple epochs, we can obtain a confident $f_b$ that overfits biased features from a smaller sample size (Figure \ref{fig:confidence}, bottom). We show in Section \ref{sec:framework} that overconfident $f_b$ is particularly important to better identify biased examples.

\section{Self-debiasing Framework}
\label{sec:framework}
We propose a self-debiasing framework that enables existing debiasing methods to work without requiring a dataset-specific knowledge about the biases' characteristics. Our framework consists of two stages: (1) automatically identifying biased examples using a shallow model; and (2) using this information to train the main model through the existing debiasing methods, which are augmented with our proposed annealing mechanism.

\subsection{Biased examples identification} First, we train a shallow model $f_b$, which approximates the behavior of a simple hand-crafted model that is commonly used by the existing debiasing methods to identify biased examples. As mentioned in Section \ref{para:bias_reliant}, we obtain $f_b$ for each task by training a copy of the main model on a small random subset of the dataset for several epochs. 
The model $f_b$ is then used to make predictions on the remaining \emph{unseen} training examples. Given a training example $\{x^{(i)}, y^{(i)}\}$, we denote the output of the shallow model as $f_b(x^{(i)}) = p_b^{(i)}$. 

Probabilities $p_b$ are assigned to each training instance to indicate how likely that it contains biases. Specifically, the presence of biases can be estimated using the scalar probability value of $p_b^{(i)}$ on the correct label, which we denote as $p_b^{(i,c)}$, where $c$ is the index of the correct label. We can interpret $p_b^{(i,c)}$ by the following: when the model predicts an example $x^{(i)}$ correctly with high confidence, i.e., $p_b^{(i,c)} \rightarrow 1$, $x^{(i)}$ is potentially biased. Conversely, when the model makes an overconfident error, i.e., $p_b^{(i,c)} \rightarrow 0$, $x^{(i)}$ is likely to be a harder example from which models should focus on learning. 

\subsection{Debiased training objective} We use the obtained $p_b$ to train the main model $f_d$ parameterized by $\theta_d$. Specifically, $p_b$ is utilized by the existing model-agnostic debiasing methods to down-weight the importance of biased examples in the training objective. In the following, we describe how the three recent model-agnostic debiasing methods (example reweighting \cite{schuster2019towards}, product-of-expert \cite{He2019UnlearnDB, Clark2019DontTT, mahabadi2019simple}, and confidence regularization \cite{Utama2020Debias}) operate within our framework:

\paragraph{Example reweighting} This method adjusts the importance of a training instance by directly assigning a scalar weight that indicates whether the instance exhibits a bias. Following \citet{Clark2019DontTT}, this weight scalar is computed as $1-p_b^{(i,c)}$. The individual loss term is thus defined as:
$$
    \mathcal{L}(\theta_d) = -(1-p_b^{(i,c)})y^{(i)} \cdot \log p_d
$$
Where $p_d$ is the softmax output of $f_d$. This formulation means that the contribution of an example to the overall loss is steadily decreased as the shallow model assigns a higher probability to the correct label (i.e., more confident prediction).

\paragraph{Product-of-expert} In this method, the main model $f_d$ is trained in an ensemble with the shallow model $f_b$, by combining their softmax outputs. The ensemble loss on each example is defined as:
$$
\mathcal{L}(\theta_d) = -y^{(i)} \cdot \log \mathtt{softmax}(\log p_d + \log p_b)
$$
During the training, we only optimize the parameters of $f_d$ while keeping the parameters of $f_b$ fixed. At test time, we use only the prediction of $f_d$.

\paragraph{Confidence regularization} This method works by regularizing model confidence on the examples that are likely to be biased. \citet{Utama2020Debias} use a self-distillation training objective \cite{Furlanello2018BornAgainNN, Hinton2015DistillingTK}, in which the supervision by the teacher model is scaled down using the output of the shallow model. The loss on each individual example is defined as a cross entropy between $p_d$ and the scaled teacher output:
$$
\mathcal{L}(\theta_d) = -\mathtt{S}(p_t, p_b^{(i,c)}) \cdot \log p_d
$$
Where $f_t$ is the teacher model (parameterized identically to $f_d$) that is trained using a standard cross entropy loss on the full dataset, and $f_t(x)=p_t$. 
This ``soft'' label supervision provided by the scaled teacher output discourages models to make overconfident predictions on examples containing biased features.

\subsection{Annealing mechanism} Our shallow model $f_b$ is likely to capture multiple types of bias, leading to more examples being down-weighted in the debiased training objectives. As a result, the effective training data size is reduced even more, which leads to a substantial in-distribution performance drop in several debiasing methods \cite{He2019UnlearnDB, Clark2019DontTT}.
To mitigate this, we propose an \emph{annealing mechanism} that allows the model to gradually learn from all examples, including ones that are detected as biased.
This is done by steadily lowering $p_b^{(i,c)}$ as the training progresses toward the end.
At training step $t$, the probability vector $p_b^{(i)}$ is scaled down by re-normalizing all probability values that have been raised to the power of $\alpha_t$: 
$
p_b^{(i,j)} = \frac{p_b^{(i,j)^{\alpha_t}}}{\sum_{k=1}^K p_b^{(i,k)^{\alpha_t}}}
$
, where $K$ is the number of labels and index $j \in \{1,...,K\}$. The value of $\alpha_t$ is gradually decreased throughout the training using a linear schedule. Namely, we set the value of $\alpha_t$ to range from the maximum value $1$ at the start of the training to the minimum value $a$ in the end of the training:
$\alpha_t = 1 - t\frac{(1-a)}{T}$, 
where $T$ is the total number of training steps. In the extreme case where $a$ is set to $0$, $p_b$ vectors are scaled down closer to uniform distribution near the end of the training. This results in a more equal importance of all examples, which is equivalent to the standard cross entropy loss. 

We note that since this mechanism gradually exposes models to potentially biased instances, it presents the risk of model picking up biases and adopting back the baseline behavior. However, our results and analysis suggest that when the parameter $a$ is set to a value close to $1$, the annealing mechanism can still provide an improvement on the in-distribution data while retaining a reasonably well performance on the challenge test sets.
\newcommand\Tstrut{\rule{0pt}{2.6ex}}         %
\newcommand\Bstrut{\rule[-0.9ex]{0pt}{0pt}}   %
\begin{table*}
    \centering 
    \footnotesize
    \setlength{\tabcolsep}{3pt}
    \begin{tabular}{r l l r | l l r | c c c r c r}
        \toprule
         \multirow{2}{*}{\textbf{Method}} & \multicolumn{3}{c|}{\textbf{MNLI (Acc.)}} & \multicolumn{3}{c|}{\textbf{FEVER (Acc.)}} & \multicolumn{6}{c}{\textbf{QQP (F1)}}\\
         & \textbf{dev} & \textit{\textbf{HANS}} & \textbf{$\Delta$} & \textbf{dev} & \textit{\textbf{symm.}} & \textbf{$\Delta$} & \textbf{D\textsubscript{ dev}} & \textbf{$\neg$D\textsubscript{ dev}} & \textit{\textbf{D\textsubscript{ PAWS}}}  & \textbf{$\Delta$} & \textit{\textbf{$\neg$D\textsubscript{ PAWS}}} & \textbf{$\Delta$} \Tstrut\\
        
        \hline
        BERT-base &  $84.5$ &  $61.5$ & - & $85.6$ & $63.1$ & - & $87.9$ & $92.9$ & $48.7$ & - & $17.6$ & -\Tstrut\Bstrut\\
        \hline
        
        Reweighting\textsubscript{ known-bias} & $83.5^{\ddag}$&  $69.2^{\ddag}$ & $+7.7$ & $84.6^{\clubsuit}$ & $66.5^{\clubsuit}$ & $\bm{+3.4}$ & $85.5$ & $91.9$ & $49.7$ & $\bm{+1.0}$ & $51.2$ & $+33.6$\Tstrut\\
        
        Reweighting\textsubscript{ \textbf{self-debias}} & $81.4$ &  $68.6$ & $+7.1$ & $87.2$ & $65.6$ & $+2.5$ & $75.7$ & $86.7$ & $43.7$ & $-5.0$ & $69.9$ & $\bm{+52.3}$\\
        
        Reweighting $\spadesuit$\textsubscript{ \textbf{self-debias}} & $82.3$ &  $69.7$ & $\bm{+8.2}$ & $87.1$ & $65.5$ & $+2.4$ & $79.4$ & $88.6$ & $46.4$ & $-2.3$ & $61.8$ & $+44.2$\Bstrut\\
        \hline
        
        PoE\textsubscript{ known-bias} & $82.9^{\ddag}$ &  $67.9^{\ddag}$ & $+6.4$ & $86.5^{\dag}$ & $66.2^{\dag}$ & $\bm{+3.1}$ & $84.3$ & $91.4$ & $50.3$ & $\bm{+1.6}$ & $61.2$ & $+43.6$\Tstrut\\
        
        PoE\textsubscript{ \textbf{self-debias}} & $80.7$ & $68.5$ & $\bm{+7.0}$ & $85.4$ & $65.3$ & $+2.1$ & $77.4$ & $87.7$ & $44.1$ & $-4.6$ & 69.4 & $\bm{+51.8}$\\
        
        PoE $\spadesuit$\textsubscript{ \textbf{self-debias}} & $81.9$ & $66.8$ & $+5.3$ & $85.9$ & $65.8$ & $+2.7$ & $80.7$ & $89.3$ & $47.4$ & $-1.3$ & $59.8$ & $+42.2$\Bstrut\\
        
        \hline
        Conf-reg\textsubscript{ known-bias} & $84.5^{\flat}$ & $69.1^{\flat}$ & $\bm{+7.6}$ & $86.4^{\flat}$ & $66.2^{\flat}$ & $\bm{+3.1}$ & $85.0$ & $91.3$ & $49.0$ & $+0.3$ & $30.9$ & $+13.3$\Tstrut\\
        
        Conf-reg\textsubscript{ \textbf{self-debias}} & $83.9$ & $67.7$ & $+6.2$ & $87.9$ & $66.1$ & $+3.0$ & $83.9$ & $90.6$ & $49.2$ & $\bm{+0.5}$ & $33.1$ & $\bm{+15.5}$\\
        
        Conf-reg $\spadesuit$\textsubscript{ \textbf{self-debias}} & $84.3$ & $67.1$ & $+5.6$ & $87.6$ & $66.0$ & $+2.9$ & $85.0$ & $91.3$ & $48.8$ & $+0.1$ & $28.7$ & $+11.1$\Bstrut\\
        \bottomrule
    \end{tabular}
    \caption{Models' performance when evaluated on MNLI, Fever, QQP, and their corresponding challenge test sets. The \texttt{known-bias} results for MNLI and FEVER are taken from \citet{Utama2020Debias}($\flat$), \citet{Clark2019DontTT}($\ddag$), \citet{mahabadi2019simple}($\dag$), and \citet{schuster2019towards}($\clubsuit$). The results of the proposed framework are indicated by \texttt{self-debias}. ($\spadesuit$) indicates the training with our proposed \emph{annealing mechanism}. Boldface numbers indicate the highest challenge test set improvement for each debiasing setup on a particular task.}
    \label{tab:all_results}
\end{table*}

\section{Experimental Setup}
\subsection{Evaluation Tasks} We perform evaluations on three NLU tasks: natural language inference, fact verification, and paraphrase identification. 
We \emph{simulate} a setting where we have not enough information about the biases for training a debiased model, and thus biased examples should be identified automatically. 
Therefore, we only use the existing challenge test set for each examined task strictly for evaluation and do not use the information about their corresponding bias types during training. 
In the following, we briefly discuss the datasets used for training on each task as well as their corresponding challenge test sets to evaluate the impact of debiasing methods:
\paragraph{Natural language inference} We use the English Multi-Genre Natural Language Inference (MNLI) dataset \cite{Williams2018ABC} which consists of 392K pairs of premise and hypothesis sentences annotated with their textual entailment information. We test NLI models on lexical overlap bias using HANS evaluation set \cite{McCoy2019RightFT}. It contains examples, in which premise and hypothesis sentences that consist of the same set of words may not hold an entailment relationship, e.g., ``cat caught a mouse'' vs. ``mouse caught a cat''. Since word overlapping is biased towards entailment in MNLI, models trained on this dataset often perform close to a random baseline on HANS.

\paragraph{Paraphrase identification} We experiment with the Quora Question Pairs dataset.\footnote{The dataset is available at \url{https://www.kaggle.com/c/quora-question-pairs}} It consists of 362K questions pairs annotated as either \textit{duplicate} or \textit{non-duplicate}. We perform an evaluation using PAWS dataset \cite{Zhang2019PAWSPA} to test whether the resulting models perform the task by relying on lexical overlap biases.

\paragraph{Fact verification} We run debiasing experiments on the FEVER dataset \cite{Thorne2018TheFE}. It contains pairs of claim and evidence sentences labeled as either \textit{support}, \textit{refutes}, and \textit{not-enough-information}. We evaluate on the FeverSymmetric test set \cite{schuster2019towards}, which is collected to reduced claim-only biases (e.g., negative phrases such as ``refused to'' or ``did not'' are associated with the \textit{refutes} label).

\subsection{Main Model} We apply our self-debiasing framework on the BERT model \cite{devlin2018bert}, which performs very well on the three considered tasks.\footnote{We use the pre-trained $\texttt{bert-base-uncased}$ model available at \url{https://huggingface.co/transformers/pretrained_models.html}.} It also shows substantial improvements on the corresponding challenge datasets when trained through the existing debiasing methods \cite{Clark2019DontTT, He2019UnlearnDB}.
For each examined debiasing method, we show the comparison between the results when it is applied within our framework and when it is trained using prior knowledge to detect training examples with a specific bias.
For the second scenario, MNLI and QQP models are debiased using a lexical overlap bias prior, whereas FEVER model is debiased using hand-crafted claim-only biased features. 
We use the results reported in their corresponding papers.
Additionally, we train a baseline BERT model with a standard cross entropy loss.

\subsection{Hyperparameters} \label{sub:hyperparam}
The hyperparameters of our framework include the number of training samples and epochs to train the shallow model $f_b$ as well as parameter $a$ to schedule the annealing process. We only use the training data, and no information about the challenging sets, for tuning these parameters.
Based on the insight from our synthetic bias analysis (Section \ref{sec:motivation}), we choose the sample size and the number of epochs which result in $f_b$ that satisfies the following conditions: (1) its accuracy on the unseen training examples is around $60\%$ to $70\%$; (2) More than $90\%$ of their predictions fall into the high confidence bin ($>0.9$). These variables vary for each task depending on their diversity and difficulty. For instance, it takes $2000$ examples and $3$ epochs of training for MNLI, and only $500$ examples and $4$ epochs for an easier task such as QQP.\footnote{We perform a search on all combinations of ${1,2,3,4,}$ and $5$ epochs and $500, 1000, 1500,$ and $2000$ examples.} 
For the annealing mechanism, we set $a=0.8$ as the minimum value of $\alpha_t$ for all experiments across the three tasks. Although this may not be an optimal configuration for all tasks, it still allows us to observe how gradually increasing the importance of ``biased'' examples may affect the overall performance.

\begin{table}
    \centering 
    \footnotesize
    \setlength{\tabcolsep}{4pt}
    \begin{tabular}{l c | l l l}
        \toprule
        \multirow{2}{*}{\textbf{Dataset}} & \multirow{2}{*}{base.} & \multicolumn{3}{c}{\textbf{confidence-regularization (\textbf{$\Delta$})}}\\
        & & known\textsubscript{ HANS} & self-deb. & self-deb. $\spadesuit$ \Tstrut\\
        
        \hline
        SICK & 55.2 & $+1.2$ \colorbox{lightgray}{$\Rightarrow$} & $+3.0$ \colorbox{lightgray}{$\Longrightarrow$} & $+2.1$ \colorbox{lightgray}{$\Longrightarrow$} \Tstrut\Bstrut\\
        
        RTE & $63.6$ & $-0.5$ \colorbox{darkgray}{\color{white}$\Leftarrow$} & $+0.5$ \colorbox{lightgray}{$\Rightarrow$} & $+0.6$ \colorbox{lightgray}{$\Rightarrow$} \Tstrut\Bstrut\\
        
        Diag. & $58.6$ & $-0.6$ \colorbox{darkgray}{\color{white}$\Leftarrow$} & $+0.4$ \colorbox{lightgray}{$\Rightarrow$} & $+0.5$ \colorbox{lightgray}{$\Rightarrow$} \Tstrut\Bstrut\\
        
        Scitail & $65.4$ & $+1.4$ \colorbox{lightgray}{$\Longrightarrow$} & $+0.4$ \colorbox{lightgray}{$\Rightarrow$} & $+1.0$ \colorbox{lightgray}{$\Longrightarrow$} \Tstrut\Bstrut\\

        \bottomrule
    \end{tabular}
    \caption{Accuracy results of self-debias confidence regularization on cross-dataset evaluation.}
    \label{tab:cross_data}
\end{table}

\section{Results and Discussion}
\paragraph{Main results} We experiment with several training methods for each task: the baseline training, debiased training with prior knowledge, and the debiased training using our self-debiasing framework (with and without annealing mechanism). We present the results on the three tasks in Table~\ref{tab:all_results}. Each model is evaluated both in terms of their in-distribution performance on the original development set and their out-of-distribution performance on the challenge test set. For each setting, we report the average results across 5 runs.

We observe that: (1) models trained through self-debiasing framework obtain equally high improvements on challenge sets of the three tasks compared to their corresponding debiased models trained with a prior knowledge (indicated as \texttt{known-bias}). In some cases, the existing debiasing methods can even be more effective when applied using the proposed framework, e.g., \texttt{self-debias} example reweighting obtains $52.3$ F1 score improvement over the baseline on the non-duplicate subset of PAWS (compared to $33.6$ by its \texttt{known-bias} counterpart). This indicates that the framework is equally effective in identifying biased examples without previously needed prior knowledge; (2) Most improvements on the challenge datasets come at the expense of the in-distribution performance (dev column) except for the confidence regularization models. For instance, the \texttt{self-debias} product-of-expert (PoE) model, without annealing, performs 2.2pp lower than the \texttt{known-bias} model on MNLI dev set. This indicates that self-debiasing may identify more potentially biased examples and thus effectively omit more training data; (3) Annealing mechanism (indicated by $\spadesuit$) is effective in mitigating this issue in most cases, e.g., improving PoE by $0.5$pp on FEVER dev and $1.2$pp on MNLI dev while keeping relatively high challenge test accuracy. Self-debias reweighting augmented with annealing mechanism even achieves the highest HANS accuracy in addition to its improved in-distribution performance.

\paragraph{Cross-datasets evaluation}
Previous work demonstrated that targeting a specific bias to optimize performance in the corresponding challenge dataset may bias the model in other unwanted directions, which proves to be counterproductive in improving the overall robustness \cite{nie2019analyzing, Teney2020OnTV}. One way to evaluate the impact of debiasing methods on the overall robustness is to train models on one dataset and evaluate them against other datasets of the same task,
which may have different types and amounts of biases \cite{belinkov-etal-2019-dont}.
A contemporary work by \citet{wu2020improving} specifically finds that 
debiasing models based on only a single bias results in models that perform significantly worse upon cross-datasets evaluation for the reading comprehension task.

Motivated by this,
we perform similar evaluations for models trained on MNLI through the three debiasing setups: \texttt{known-bias} to target the HANS-specific bias, \texttt{self-debiasing}, and \texttt{self-debiasing} augmented with the proposed annealing mechanism. We do not tune the hyperparameters for each target dataset and use the models that we previously reported in the main results. As the target datasets, we use 4 NLI datasets: Scitail \cite{khotSC18}, SICK \cite{marelli-etal-2014-sick}, GLUE diagnostic set \cite{wang2018glue}, and 3-way version of RTE 1, 2, and 3 \cite{daganRte1, barHaimRte2, giampiccolo-etal-2007-rte3}.\footnote{We compiled and reformated the dataset files which are available at \url{https://nlp.stanford.edu/projects/contradiction/}.}

We present the results in Table \ref{tab:cross_data}. We observe that the debiasing with prior knowledge to target the specific lexical overlap bias (indicated by \texttt{known\textsubscript{HANS}}) can help models to perform better on SICK and Scitail.
However, its resulting models under-perform the baseline in RTE sets and GLUE diagnostic, degrading the accuracy by $0.5$ and $0.6$pp. In contrast, the self-debiased models, with and without annealing mechanism, outperform the baseline on all target datasets, both achieving additional $1.1$pp on average. 
The gains by the two self-debiasing suggest that while they are effective in mitigating the effect of one particular bias (i.e., lexical overlap), they do not result in models learning other unwanted patterns that may hurt the performance on other datasets.
These results also extend the findings of \citet{wu2020improving} to the NLU settings in that addressing multiple biases at once, as done by our general debiasing method, leads to a better overall generalization.

\begin{figure}
    \centering
    \includegraphics[width=0.9\columnwidth]{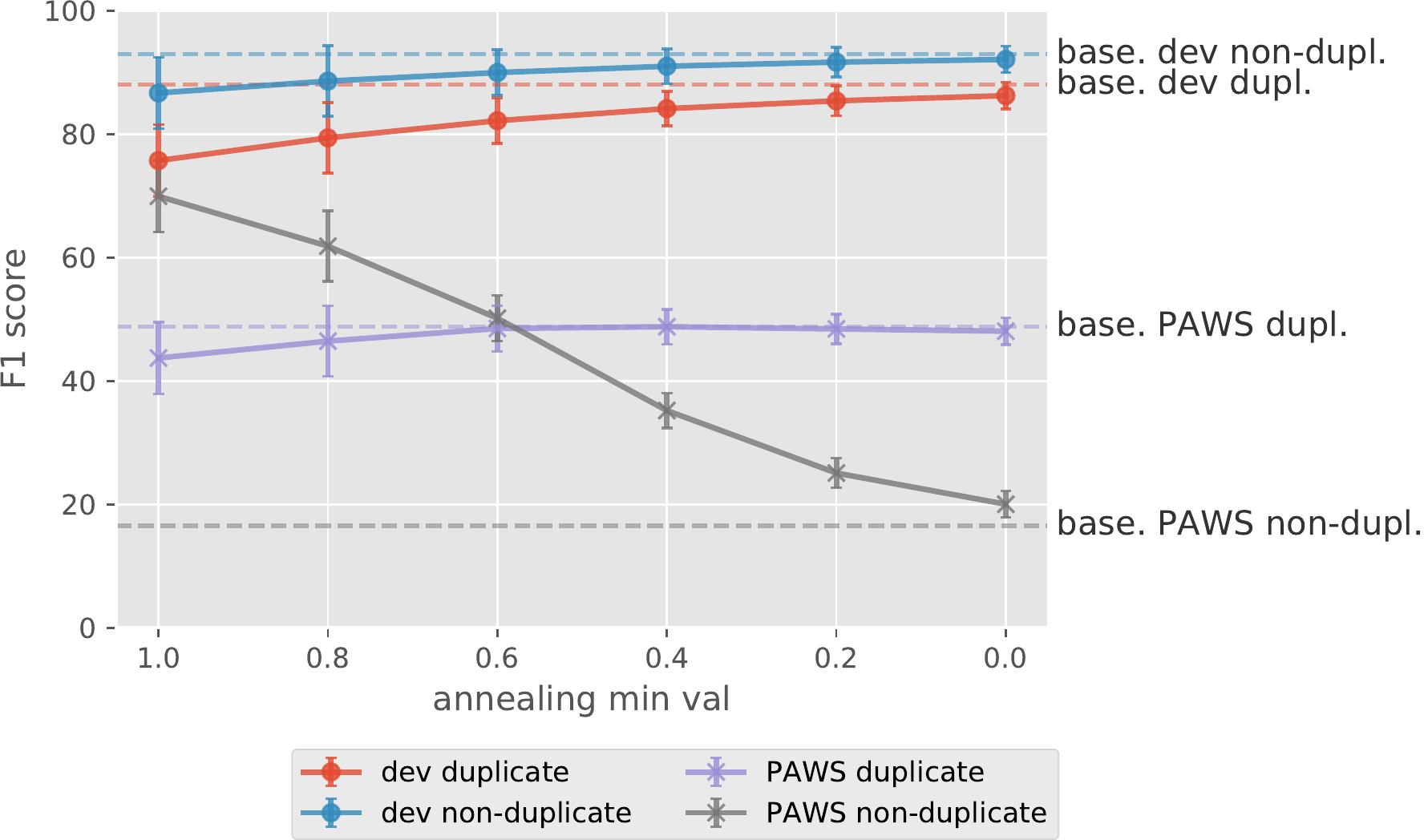}
    \caption{Analysis of the annealing mechanism using different values of minimum $\alpha_t$.}
    \label{fig:annealing}
\end{figure}

\paragraph{Analyzing the annealing mechanism} In previous experiments, we show that setting the minimum $\alpha_t$ to only slightly lower than $1$ (i.e., $a=0.8$) results in improvements on the in-distribution without substantial degradation on challenge datasets scores. We question whether this behavior persists once we set $a$ closer to $0$. Specifically, do models fall back to the baseline performance when the loss gets more equivalent to the standard cross-entropy at the end of the training?

We run additional experiments using the self-debiased example reweighting on QQP $\Rightarrow$ PAWS evaluations. We consider the following values to set the minimum $\alpha_t$: $1.0, 0.8, 0.6, 0.4, 0.2,$ and $0.0$. For each experiment, we report the average scores across multiple runs. As we see in Figure \ref{fig:annealing}, the challenge test scores decrease as we set minimum $a$ to lower values. Annealing can still offer a reasonable trade-off between in-distribution and challenge test performances up until $a=0.6$, before falling back to baseline performance at $a=0$.
These results suggest that models are still likely to learn spurious shortcuts from biased examples that they are exposed to even at the end of the training. Consequently, the annealing mechanism should be used cautiously by setting the minimum $\alpha_t$ to moderate values, e.g., 0.6 or 0.8.

\begin{figure}
    \centering
    \includegraphics[width=.8\columnwidth]{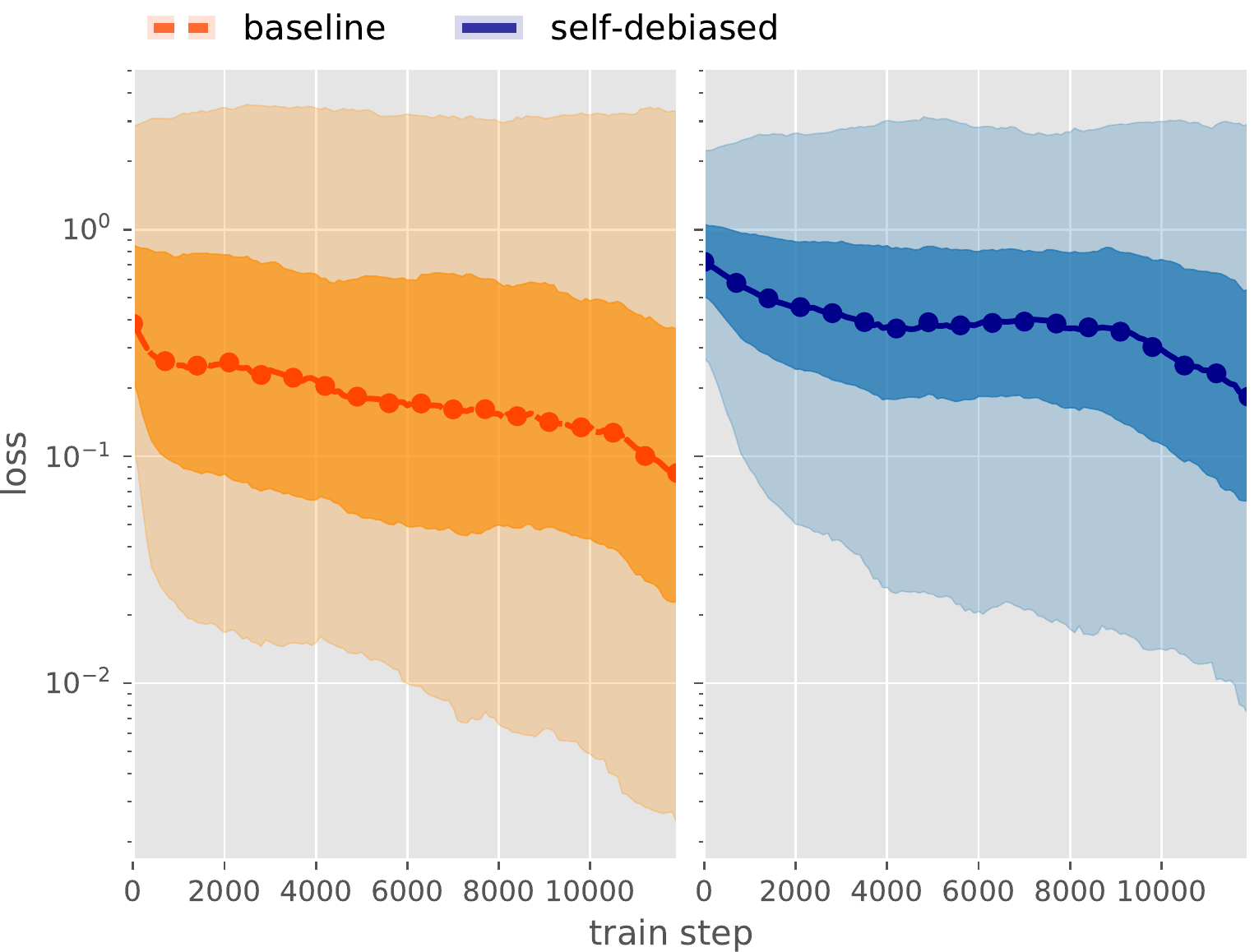}
    \caption{Training loss curves for the first 15K steps by the baseline and self-debias example reweighting training (shown in log scale). Solid lines indicate the median loss within each training batch. The dark and light shadow areas represent the range between 25th to 75th percentile and the range between 0th (minimum) and 100th percentile (maximum), respectively.}
    \label{fig:loss_var}
\end{figure}

\paragraph{Impact on learning dynamics} 
We previously show (Figure \ref{fig:learning_curve}) that baseline models tend to learn easier examples more rapidly, allowing them to make correct predictions by relying on biases. As the self-debiasing framework manages to mitigate this fallible reliance, we expect some changes in models' learning dynamics.
We are, therefore, interested in characterizing these changes by analyzing their training loss curve. In particular, we examine the individual losses on each training batch and measure their variability using percentiles (i.e., 0th, 25th, 50th, 75th, and 100th percentile). Figure~\ref{fig:loss_var} shows the comparison of the individual loss variability between the baseline and the self-debiased models when trained on MNLI. We observe that the median loss of the baseline model converges faster than the self-debiased counterpart (dotted solid lines). However, examples below its 25th percentile already have losses smaller than $10^{-1}$ when the majority of the losses are still high (darker shadow area). This indicates that unregularized training optimizes faster on certain examples, possibly due to the presence of biases. On the contrary, self-debiased training maintains relatively less variability of losses throughout the training. This result suggests that overconfident predictions (unusually low loss examples) can be an indication of the model utilizing biases. This is in line with the finding of \citet{Utama2020Debias}, which shows that regularizing confidence on biased examples leads to improved robustness against biases.

\paragraph{Bias identification stability} Researchers have recently observed  large variability in the generalization performance of fine-tuned BERT model  \cite{mosbach2020stability, zhang2020revisiting}, especially in the out-of-distribution evaluation settings \cite{mccoy2019berts, zhou2020curse}. This may raise concerns on whether our shallow models, which are trained on the sub-sample of the training data, can consistently learn to rely mostly on biases.
We, therefore, train 10 instances of shallow models on the MNLI dataset using different random seeds (for classifier's weight initialization and training sub-sampling). For evaluation, we perform two different partitionings of MNLI dev set based on the output of two simple hand-crafted models, which use lexical overlap and hypothesis-only features \cite{Gururangan2018AnnotationAI}, respectively. The stability of bias utilization across the runs is evaluated by measuring their performance on \textit{easy} and \textit{hard} subsets of each partitioning, where examples that simple models predicted correctly belong to \textit{easy} and the rest belong to \textit{hard}.\footnote{Although this may seem to be against the spirit of not using prior knowledge about the biases, we believe that this step is necessary to show the stability of the shallow models and to validate if they indeed capture the intended biases.}

\begin{figure}
    \centering
    \includegraphics[width=.8\columnwidth]{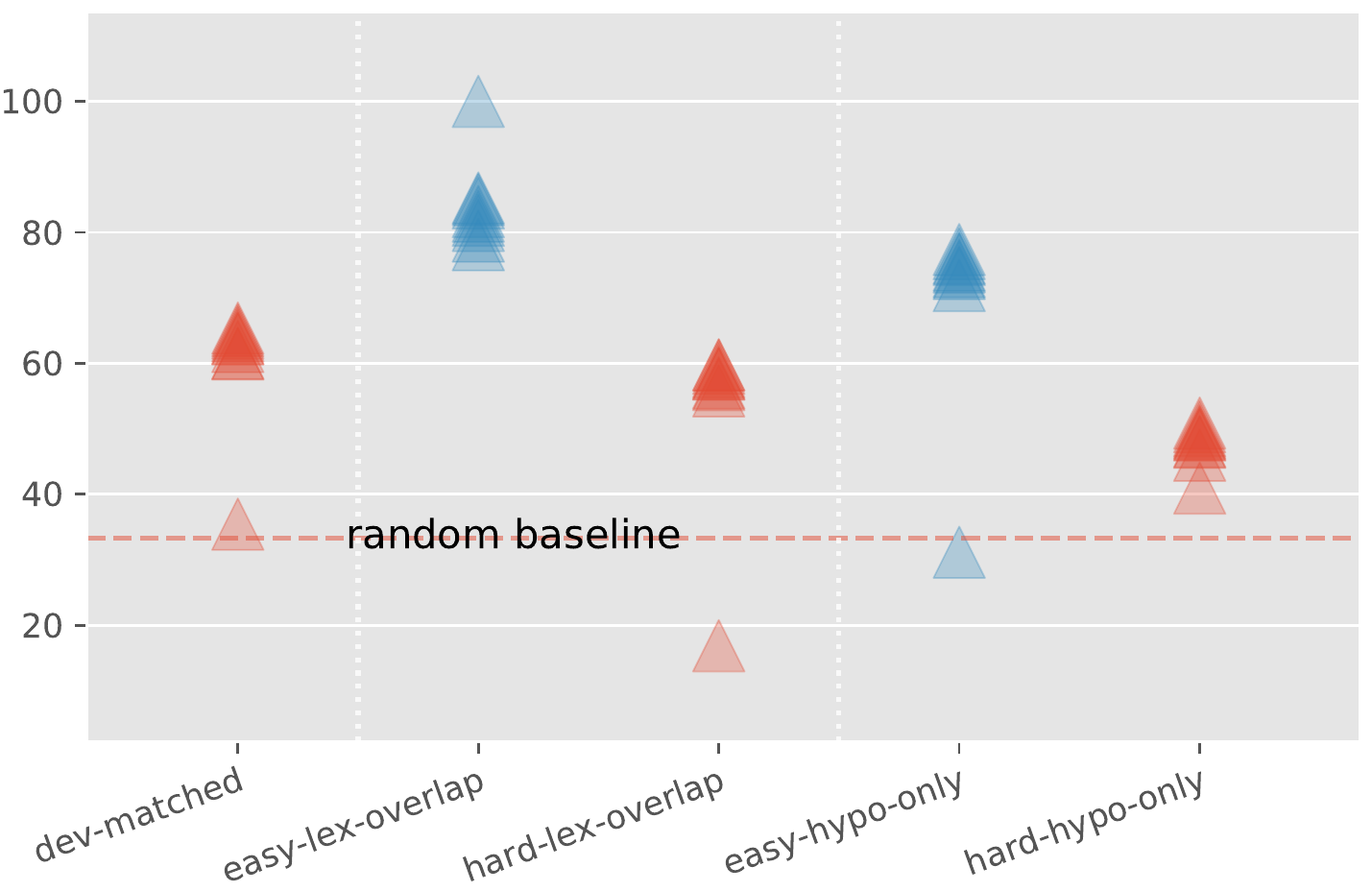}
    \caption{Evaluation of 10 shallow model instances on easy/hard partitioning of MNLI dev based on the presence of lexical overlap and hypothesis-only biases. The results suggest the stability of shallow models in capturing the two biases.}
    \label{fig:stability}
\end{figure}

Figure \ref{fig:stability} shows the results. We observe small variability in the overall dev set performance which ranges in 61-65$\%$ accuracy. Similarly, the models obtain consistently higher accuracy on the easy subsets over the hard ones: 79-85$\%$ vs. 56-59$\%$ on the lexical-overlap partitioning and 72-77$\%$ vs. 48-50$\%$ on the hypothesis-only partitioning. The results indicate that: 1) the bias-reliant behavior of shallow models is stable; and 2) shallow models capture multiple types of bias. However, we also observe one rare instance of the shallow model that fails to converge during training and is stuck at making random predictions (33$\%$ in MNLI). This may indicate that the biased examples are under-sampled in that particular run. In that case, we can easily spot this undesired behavior, discard the model, and perform another sampling.

\section{Related Work}
The artifacts of large scale dataset collections result in dataset biases that allow models to perform well without learning the intended reasoning skills. In NLI, models can perform better than chance by only using the partial input \cite{Gururangan2018AnnotationAI, Poliak2018HypothesisOB, tsuchiya-2018-performance} or by basing their predictions on whether the inputs are highly overlapped \cite{McCoy2019RightFT, Dasgupta2018EvaluatingCI}. Similar phenomena exist in various tasks, including argumentation mining \cite{niven2019probing}, reading comprehension \cite{kaushik-lipton-2018-much}, or story cloze completion \cite{schwartz-etal-2017-effect, cai-etal-2017-pay}. To allow a better evaluation of models' reasoning capabilities, researchers constructed challenge test sets composed of ``counterexamples'' to the spurious shortcuts that models may adopt \cite{Jia2017AdversarialEF, Glockner2018BreakingNS, Zhang2019PAWSPA, Naik2018StressTE}. Models evaluated on these sets often fall back to random baseline performance.

There has been a flurry of work in dynamic dataset construction to systematically reduce dataset biases through adversarial filtering \cite{Zellers2018SWAGAL, Sakaguchi2019WINOGRANDEAA, Bras2020AdversarialFO} or human in the loop \cite{Nie2019AdversarialNA, Kaushik2020Learning, gardner2020evaluating}. While promising, researchers also show that newly constructed datasets may not be fully free of hidden biased patterns \cite{sharma-etal-2018-tackling}. It is thus crucial to complement the data collection efforts with learning algorithms that are more robust to biases, such as the recently proposed product-of-expert \cite{Clark2019DontTT, He2019UnlearnDB, mahabadi2019simple}, confidence regularization \cite{Utama2020Debias}, or other training strategies \cite{belinkovAdv2019, yaghoobzadeh2019robust, tu20tacl}. Despite their effectiveness, these methods are limited by their assumption on the availability of information about the task-specific biases. Our framework aims to alleviate this limitation and enable them to address \emph{unknown} biases.

\section{Conclusion}
We present a general self-debiasing framework to address the impact of unknown dataset biases by omitting the need for thorough dataset-specific analysis to discover the types of biases for each new dataset. We adopt the existing debiasing methods into our framework and enable them to obtain equally high improvements on several \emph{challenge test sets} without targeting a specific bias. The evaluation also suggests that our framework results in better overall robustness compared to the bias-specific counterparts. Based on our analysis, future work in the direction of automatic bias mitigation may include identifying potentially biased examples in an \emph{online} fashion and discouraging models from exploiting them throughout the training.

\section*{Acknowledgments}
We thank Max Glockner, Mingzhu Wu, Johannes Daxenberger, and the anonymous reviewers for their constructive feedback. This work is funded by the German Research Foundation through the research training group AIPHES (GRK 1994/1) and  by the German Federal Ministry of Education and Research and
the Hessen State Ministry for Higher Education, Research and the Arts within their joint support of the National Research Center for Applied Cybersecurity ATHENE.

\bibliography{emnlp2020}
\bibliographystyle{acl_natbib}

\appendix
\clearpage
\section{Natural Language Inference}
\paragraph{Main model} We finetune the BERT base model for all settings (baseline, known-bias, and self-debiasing) using default parameters: 3 epochs of training with learning rate $5^{-5}$. An exception is made for product-of-expert and confidence regularization, where we follow \citet{He2019UnlearnDB} to run the training longer, i.e. 5 epochs.
\paragraph{Shallow model} The shallow model for MNLI is trained on 2K of examples for 3 epochs using the default learning rate of $5^{-5}$.

\section{Fact verification} 
\paragraph{Main model} We follow \citet{schuster2019towards} in finetuning the BERT base model on FEVER dataset using the following parameters: learning rate $2^{-5}$ and 3 epochs of training.

\paragraph{Shallow model} The shallow model can be trained in lesser amount of data, 500 examples. We train the model for 5 epochs with the same learning rate, $2^{-5}$.

\section{Paraphrase Identification}
\paragraph{Main model} We follow \citet{Utama2020Debias} in setting the parameters for training a QQP model: learning rate $2^{-5}$ and 3 epochs of training.

\paragraph{Shallow model} Similar to FEVER, we train the shallow model using only 500 examples. It converges in 4 epochs using the same learning rate, $2^{-5}$.

\section{Synthetic MNLI Results} We report the final accuracy of models when trained on our synthetic bias datasets. We show that the anti-biased accuracy correlates negatively with the proportion of the biased examples. We present the results in Table \ref{tab:synthetic_bias_results}.

\begin{table}[H]
    \centering 
    \footnotesize
    \setlength{\tabcolsep}{4pt}
    \begin{tabular}{r | l l l}
        \toprule
        \multirow{2}{*}{\textbf{Bias prop.}} & \multicolumn{3}{c}{\textbf{test sets}}\\
        & original & biased & anti-biased \Tstrut\\
        
        \hline
        0.9 & $83.6$ \colorbox{lightgray}{$\Leftarrow$} & $97.1$ \colorbox{lightgray}{$\Longrightarrow$} & $61.7$ \colorbox{lightgray}{$\Longleftarrow$} \Tstrut\Bstrut\\
        
        \hline
        0.8 & $83.7$ \colorbox{lightgray}{$\Leftarrow$} & $95.3$ \colorbox{lightgray}{$\Longrightarrow$} & $70.4$ \colorbox{lightgray}{$\Longleftarrow$} \Tstrut\Bstrut\\
        
        \hline
        0.7 & $83.9$ \colorbox{lightgray}{$\Leftarrow$} & $92.8$ \colorbox{lightgray}{$\Rightarrow$} & $75.5$ \colorbox{lightgray}{$\Leftarrow$} \Tstrut\Bstrut\\
        
        \hline
        0.6 & $84.1$ \colorbox{lightgray}{$=$} & $90.9$ \colorbox{lightgray}{$\Rightarrow$} & $78.5$ \colorbox{lightgray}{$\Leftarrow$} \Tstrut\Bstrut\\
        
        \bottomrule
    \end{tabular}
    \caption{Final accuracy of models trained on synthetic bias datasets.}
    \label{tab:synthetic_bias_results}
\end{table}

\section{Detailed HANS Results}
HANS dataset \cite{McCoy2019RightFT} consist of three subsets, covering different inference phenomena which happen to have lexical overlap: (a) {Lexical overlap} e.g., ``\textit{{The doctor} was {paid} by {the actor}}'' vs. ``\textit{The doctor paid the actor}''; (b) {Subsequence}, e.g., ``\textit{The doctor near {the actor danced}}'' vs. ``\textit{The actor danced}''; and (c) {Constituent} e.g., ``\textit{If {the artist slept}, the actor ran}'' vs. ``\textit{The artist slept}''. Each subset contains examples of both entailment and non-entailment. The 3-way predictions on MNLI is mapped to HANS by taking max pool between neutral and contradiction labels. We present the results of our experiments in Table \ref{tab:hans_all}.

\begin{table}[H]
    \centering 
    \footnotesize
    \setlength{\tabcolsep}{3pt}
    \begin{tabular}{r l l l l l l}
        \toprule
         \multirow{2}{*}{\textbf{Method}} & \multicolumn{6}{c}{\textbf{HANS all sets (Acc.)}}\\
         & \textbf{Lex} & \textbf{Lex.} & \textbf{Sub.} & \textbf{Sub.} & \textbf{Con.} & $\neg$\textbf{Con.}\\
        
        \midrule
        BERT-base & 96.0 & 51.8 & 99.5 & 7.4 & 99.4 & 14.5 \\
        \midrule
        
        Rew.\textsubscript{ \textbf{self-debias}} & 81.3 & 73.3 & 94.7 & 34.5 & 92.8 & 42.3\\
        
        Rew. $\spadesuit$\textsubscript{ \textbf{self-debias}} & 84.7 & 77.1 & 96.0 & 30.5 & 95.3 & 37.4 \\
        \midrule
        
        PoE\textsubscript{ \textbf{self-debias}} & 77.0 & 73.6 & 92.1 & 42.2 & 89.3 & 49.8 \\
        
        PoE $\spadesuit$\textsubscript{ \textbf{self-debias}} & 78.5 & 67.7 & 91.3 & 28.6 & 95.4 & 45.1 \\
        
        \midrule
        
        Conf-reg\textsubscript{ \textbf{self-debias}} & 81.8 & 78.2 & 93.7 & 31.7 & 95.1 & 31.5 \\
        
        Conf-reg $\spadesuit$\textsubscript{ \textbf{self-debias}} & 87.4 & 74.5 & 96.3 & 27.4 & 95.1 & 26.6 \\
        \bottomrule
    \end{tabular}
    \caption{Models' performance on HANS challenge test set \cite{McCoy2019RightFT}. Column \texttt{lex.}, \texttt{con.}, and \texttt{sub.} stand for lexical overlap, constituency, and subsequence, respectively. The ($\neg$) symbol indicates the non-entailment subset.}
    \label{tab:hans_all}
\end{table}

\label{sec:appendix}

\end{document}